\pdfoutput=1

\documentclass[11pt]{article}

\usepackage[final]{acl}
\usepackage{url}
\usepackage{times}
\usepackage{latexsym}
\usepackage{graphicx}
\usepackage{multirow}

\usepackage[T1]{fontenc}

\usepackage[utf8]{inputenc}

\usepackage{microtype}

\usepackage{inconsolata}
\usepackage{xspace}
\usepackage{booktabs}

\newcommand{\algname}{\textsc{TS-DShapley}\xspace}
\title{Data Selection for Fine-tuning Large Language Models\\ Using Transferred Shapley Values}

\author{
Stephanie Schoch$^{ }$\quad Ritwick Mishra$^{ }$\quad Yangfeng Ji$^{ }$\\
$^{ }$ Department of Computer Science\\University of Virginia\\ Charlottesville, VA 22904\\
\texttt{\{sns2gr,mbc7bu,yangfeng\}@virginia.edu}\\
}

\begin{document}
\maketitle

\begin{abstract}
Although Shapley values have been shown to be highly effective for identifying harmful training instances, dataset size and model complexity constraints limit the ability to apply Shapley-based data valuation to fine-tuning large pre-trained language models.
To address this, we propose \algname, an algorithm that reduces computational cost of Shapley-based data valuation through:
1) an efficient
sampling-based method that aggregates Shapley values computed from subsets for valuation of the entire training set,
and 2) a value transfer method that leverages value information extracted from a simple classifier trained using representations from the target language model.
Our experiments applying \algname to select data for fine-tuning BERT-based language models on benchmark natural language understanding (NLU) datasets 
show that \algname outperforms existing data selection methods. Further, \algname can filter fine-tuning data to increase language model performance compared to training with the full fine-tuning dataset.
\end{abstract}

\section{Introduction}\label{sec:intro}
Large language models (LMs) have achieved state-of-the-art performance on many natural language processing (NLP) tasks \citep{radford2019language, brown2020language, sanh2022multitask}. To adapt these models to new datasets and tasks, the standard approach is to fine-tune a pre-trained LM on a targeted downstream task. This allows the pre-trained general linguistic knowledge to be leveraged while fine-tuning to learn the task-specific information. However, during fine-tuning, pre-trained LMs are prone to significant performance degradation in the presence of noisy data \citep{srivastava2020noisy}. This effect may be further amplified when noisy or otherwise harmful instances are highly influential to the model parameters \citep{koh2017understanding}. As a result, it is important to identify harmful instances in the fine-tuning data that may obfuscate the task information and degrade performance.

\begin{figure}[t!]
\includegraphics[width=\linewidth]{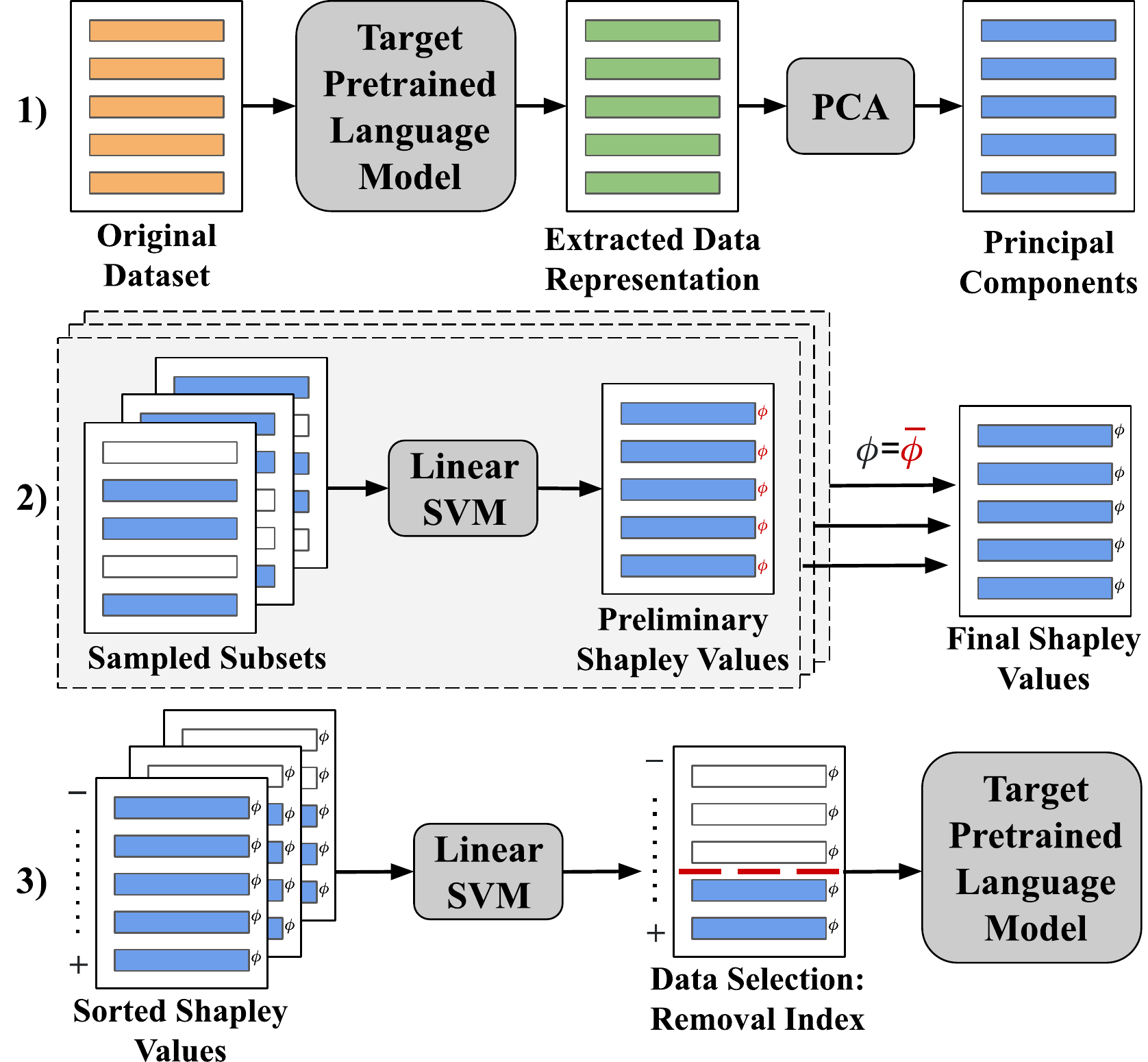}
\caption{\label{fig:pipeline} An overview of \algname: 1) Process the data using the target LM; 2) Compute \textit{sampling} chains using a subset of the training set
and aggregate the resulting Shapley values; and 3) \textit{Transfer} the estimated data value information for use with the target LM by estimating the optimal low value data removal index.}
\end{figure}

To automatically identify harmful data, prior works have used training dynamics \citep{swayamdipta2020dataset} and estimation of marginal contributions via leave-one-out retraining \citep{cook1977detection} or influence functions \citep{koh2017understanding}. Shapley values, which satisfy certain desirable fairness guarantees, have also recently been adopted from cooperative game theory to measure datum contributions, where a data point's Shapley value is the average marginal contribution to every possible data subset \citep{ghorbani2019data}.

In practice, Shapley-based data values are approximated using various techniques \citep{ghorbani2019data, jia2019towards, jia2021scalability, kwon2021beta, schochcs}, as exact Shapley value computation over a dataset would require \textit{exhaustively retraining the model} for every datum on every possible subset (i.e. \textit{exponential complexity with respect to the number of data points}). However, many of the existing approximation methods still exhibit a computational bottleneck when considering datasets and models at scale (e.g. datasets larger than 5K instances).
This, in turn, directly limits the application of Shapley-based data valuation to state-of-the-art LMs and many NLP datasets.

To address the challenges posed by 1) the \textit{model constraint} (the model retraining requirement) and 2) the \textit{dataset constraint} (the time-complexity/dataset size relation), we propose Transferred Sampling Data Shapley (\algname), an algorithm that utilizes two novel components that directly address each constraint. Specifically, to address the model constraint, we propose to compute Shapley-based data values using a simple, linear model that is trained on the learned representation from the target LM. Additionally, to address the dataset constraint, we propose a sampling-based method that computes Shapley values on data subsets and aggregates them for valuation of the entire training set.

Our contributions are as follows: 1) we propose a sampling-based data Shapley computation method and demonstrate its efficacy empirically using as little as $2\%$ of the original training data; 2) we propose the use of a simple linear classifier with a target model's pre-trained representation and demonstrate empirically the performance gains achieved over alternate pre-trained embeddings; 
and 3) we show the efficacy of Shapley-based data valuation and selection methods on benchmark NLU tasks using fine-tuned large LMs.\footnote{Code is available at \url{https://github.com/stephanieschoch/ts-dshapley}}

\section{Related Work}\label{sec:related}
While Shapley values are often applied in a post hoc manner following model training \citep{ghorbani2019data, kwon2021beta, jia2019efficient, jia2019towards, jia2021scalability, schochcs}, the demonstrated efficacy 
makes it a natural extension to apply such methods for data selection \textit{prior to} training. 
To this end, Shapley values have been used for evaluating data for transfer learning \citep{parvez2021evaluating} and in active learning \citep{ghorbani2021data}. 

Further, although Shapley-based data values have primarily been considered model-specific, in practice, a subset of training instances that may harm performance may be mislabeled \citep{koh2017understanding, swayamdipta2020dataset, ghorbani2019data} or exhibit spelling mistakes or grammatical errors \citep{sun2020adv, srivastava2020noisy}, which should be intrinsic to the dataset.
Prior works have demonstrated the transferability of Shapley-based data values across various classifier architectures \citep{schochcs} and have demonstrated the efficacy of surrogate KNN classifiers using pre-trained embeddings \citep{jia2021scalability}.
Notably, our work differs in that we utilize the pre-trained embeddings extracted from the target LM and avoid the $k$-nearest neighbor assumption that
training data far from a test datum do not contribute to its prediction \citep{jia2019efficient}. 

\section{Method}\label{sec:method}
Let $D=\{(x_i, y_i)\}_{i=1}^n$ denote a training set containing $n$ training instances. 
For each training instance $i$, the Shapley value $\phi_i$ is defined as the average marginal contribution of $i$ to every possible subset $S \subseteq D$ that contains this instance 
\citep{ghorbani2019data}: 
${\small
\phi_i = \sum_{S\subseteq D; i\in S} \frac{1}{{n-1 \choose |S\backslash\{i\}|}}\{v_{\mathcal{A}}(S) - v_{\mathcal{A}}(S\backslash\{i\})\}
}$
where $v_{\mathcal{A}}(S)$ is a value function, typically defined as the development accuracy of model $\mathcal{A}$ trained on $S$.
The challenge of calculating $\phi_i$ is two-fold: the exponential complexity of all possible subsets $S\subseteq D$ and the computational cost of training $\mathcal{A}$ on each $S$ and $S\backslash\{i\}$. While Shapley-based data values are approximated in practice, most existing approximation methods are not efficient enough for large scale learning problems.

\subsection{\algname}
Let $\mathcal{A}_{tgt}$ be the target classifier (i.e. large LM) that we want to fine-tune on a subset of $D$. 
To reduce computational cost, we propose to (1) use a 
linear classifier $\mathcal{A}_{src}$ as the proxy of $\mathcal{A}_{tgt}$ for data valuation; (2) use multi-chain Monte Carlo sampling to compute Shapley values on different subsets of $D$.
For faithful data valuation, we further propose to train $\mathcal{A}_{src}$ on the data representations extracted from $\mathcal{A}_{tgt}$. 

\paragraph{Representation Extraction.} 
We extract the representations from the penultimate layer of the pre-trained LM $\mathcal{A}_{tgt}$ as the inputs for training $\mathcal{A}_{src}$.
Note that training $\mathcal{A}_{src}$ in this way is equivalent to fixing the LM and only fine-tuning the last classification layer. 
To further remove the redundancy in the representations and reduce computational cost, we follow prior work by performing PCA on the collection of representations and selecting the first 32 principal components \citep{ghorbani2019data, kwon2021beta, schochcs}.

\paragraph{Sampling Data Shapley.}
Instead of directly estimating Shapley-based data values via Monte Carlo sampling on the whole training set, our approach performs Monte Carlo sampling on subsets of the data, which we refer to as \textit{sampling chains}. Within a single sampling chain $c$, we sample a subset of training instances $S_t$, estimate their contributions, and repeat $T$ times.
The contribution of each instance in $S_t$ is calculated by removing one instance at a time in a random order. For example, 
the contribution of the first randomly removed instance $i$ is $c_{S_t}(i) = v_{\mathcal{A}_{src}}(S_t) - v_{\mathcal{A}_{src}}(S_t\backslash\{i\})$, the contribution of the second randomly removed instance $k$ is $c_{S_t}(k) = v_{\mathcal{A}_{src}}(S_t\backslash\{i\}) - v_{\mathcal{A}_{src}}(S_t\backslash\{i, k\})$, and so on. On the other hand, if an instance $i$ is not in $S_t$, $c_{S_t}(i) = 0$.

After $T$ times, the Shapley value of instance $i$ is approximated as $\phi_i\approx \frac{1}{T}\sum_{S_t}c_{S_t}(i)$.
To balance the computational efficiency and approximation, we empirically define a range of the size $|S_t|\in [\frac{s}{2}, s]$, with subset size $s$ as the sampling upper bound.

Computation can be further sped up with multiple Monte Carlo sampling chains $S_t^{(c)}, c \in \{1,\dots,J\}$. The corresponding value approximation is defined as $\phi_i = \frac{1}{J}\sum_{c}\frac{1}{T}\sum_{S^{(c)}_t}c_{S^{(c)}_t}(i).$
As each chain can be computed independently, the efficiency can be boosted with parallel computing. 
This novel idea of multi-chain sampling serves as the core of \algname and significantly speeds up computation, in practice working with 
a simple model $\mathcal{A}_{src}$.

\paragraph{Data Selection with \algname Values.} To identify harmful data points, we use the data removal strategy of \citet{ghorbani2019data} on $\mathcal{A}_{src}$ and transfer the selection outcome to the target model $\mathcal{A}_{tgt}$.
Specifically, we gradually remove training instances from the lowest estimated contribution value to the highest estimated contribution value.
Following each removal, we retrain $A_{src}$ and evaluate predictive performance on the held-out development data.
As a result, this removal procedure will identify a optimal subset $S_{opt}$ that gives the best predictive performance on $A_{src}$. 
With the assumption of data value transferability \citep{schochcs}, we expect that $\mathcal{A}_{tgt}$ trained on $S_{opt}$ will give no worse, and likely better performance, than $\mathcal{A}_{tgt}$ trained on $D$. 
While this data removal strategy is proposed in prior work \citep{ghorbani2019data}, the data selection use case is novel in NLP. 

\section{Experiments}\label{sec:exp}
\subsection{Experiment Setup}
\paragraph{Pre-trained Large Language Models.} We utilize two transformer-based large LMs for which traditional Shapley-based data value computation would be intractable: RoBERTa-base \citep[125M parameters]{liu2019roberta} and DistilBERT  \citep[66M parameters]{sanh2019distilbert}. 

\paragraph{Datasets.} 
We select one GLUE benchmark \citep{wang2019glue} dataset from each task category: SST-2 \citep{socher2013recursive}, QQP \citep{qqpdataset}, and RTE \citep{dagan2006pascal}, 
representing Single-Sentence Tasks, Similarity and Paraphrase Tasks, and Inference Tasks, respectively. Additional dataset details are reported in \autoref{app:exp-details}. Notably, we select datasets of varied sizes to reflect diverse sampling subset to training set size ratios. 

\paragraph{Data Selection Baselines.} We compare against performance when training on the full data subset as well as three selection baselines: leave-one-out (LOO) \citep{cook1977detection}, KNN-shapley (KNN) \citep{jia2019efficient, jia2021scalability}, and random sampling. For LOO, we use the same classifier architecture as with \algname to compute value estimates.
For both LOO and KNN, we reduce the dataset using the data removal procedure defined in \autoref{sec:method}.
Finally, for random sampling, we remove a random sample of data points equal to the number of points removed via \algname.

\begin{table*}[ht!]
\centering
\small
\begin{tabular}{lllllllll}%
  \toprule
   \multirow{2}{*}{\textbf{Method Category}} & \multirow{2}{*}{\textbf{Method}} & \multicolumn{3}{c}{\textbf{RoBERTa}} & & \multicolumn{3}{c}{\textbf{DistilBERT}}\\
  \cmidrule{3-5}
  \cmidrule{7-9}
  & & \textbf{SST-2} & \textbf{QQP} & \textbf{RTE} & & \textbf{SST-2} & \textbf{QQP} & \textbf{RTE} \\
  \midrule
  \multirow{3}{*}{Full Training Set} & \citet{liu2019roberta} & 0.948 & 0.919 & 0.787 & & -- & -- & --\\
  & \citet{sanh2019distilbert} & -- & -- & -- & & 0.913 & 0.885 & 0.599 \\
  & Full Dataset & 0.950 & 0.917 & 0.788 & & 0.908 & 0.905 & 0.618\\
  \midrule
  \multirow{3}{*}{Data Selection Baselines} & Leave-One-Out & 0.947 & -- & 0.784 & & 0.912 & -- & 0.614\\
  & KNN Shapley & 0.946 & 0.916 & 0.781 & & 0.911 & 0.905 & 0.622\\
  & Random & 0.947 & 0.917 & 0.684 & & 0.911 & 0.905 & 0.589\\
  \midrule
  Our Method & \algname & \textbf{0.953} & \textbf{0.919} & \textbf{0.801} & & \textbf{0.915} & \textbf{0.907} & \textbf{0.652}\\
  \bottomrule
\end{tabular}
\caption{\label{selection-table} Predictive accuracy when selecting data using each valuation method. Results reflect the mean of five trials. We do not report LOO as a baseline for QQP due to computational intractability.}
\end{table*}

\subsection{Data Selection Experiment}\label{subsec:data-selection-exp}
To test the efficacy of using \algname to select data for fine-tuning large LMs, we compute data values using each method and perform the data removal procedure described in \autoref{sec:method}. 
Specifically, we remove the lowest value data points preceding the data removal step that achieved the highest development accuracy using $A_{src}$. For \algname, we vary the subset size and number of chains based on dataset size, using subset size $= 6.7k (10\%), 7.28k (2\%), 374 (15\%)$ and number of chains $= 25, 10, 25$ for SST-2, QQP, and RTE, respectively. Additional training and hyperparameter details, including details of a limited hyperparameter sweep, can be found in \autoref{app:exp-details}.

\paragraph{Results.}
Results are shown in \autoref{selection-table}. \algname consistently outperforms baseline selection methods as well as performance using the full fine-tuning dataset. Notably, data selection using \algname resulted in performance improvements of up to $1.3\%$ and $3.4\%$ for RoBERTa and DistilBERT, respectively, over the predictive performance when training using the full fine-tuning dataset. These results indicate \algname successfully identifies data points that harm model performance. As an additional analysis, for the RTE dataset we show the location of harmful points identified by \algname on a data map \citep{swayamdipta2020dataset} in \autoref{app:results}.

\subsection{Sampling Hyperparameter Analysis}
\algname exhibited good performance for data selection across various subset sizes and numbers of chains. For example, on QQP \algname outperformed the full dataset and baseline methods when using a subset of just 2\% of the training set. 
To better understand the impact of different parameter values, we utilize a parameter value grid on the RTE dataset and re-compute \algname. Specifically, using the best hyperparameters from \autoref{subsec:data-selection-exp} (see \autoref{app:exp-details}), we evaluate performance of RoBERTa and DistilBERT using a parameter sweep of subset size as a percentage of the total training set size, subset size $ \in \{1, 2, 5, 10, 15\}\%$, and number of chains $ \in \{2, 5, 10, 15\}$ and report the Pearson's correlation between each parameter and performance. 

\paragraph{Results.}
All correlations are reported in \autoref{app:results} and summarized here. When subset size~$> 2\%$, both models demonstrate a high positive correlation between number of chains and performance. For example, when using~$15\%$ of the training data, RoBERTa on RTE had a correlation of~$0.94$. Across the different number of chains, however, there was no consistent pattern of correlation between subset size and performance. This indicates that increasing number of chains (which can be computed in parallel) may be of more benefit compared to increasing sampling subset size.

\begin{table}[t!]
\centering
\small
\begin{tabular}{lllll}%
  \toprule
  \textbf{Model} & \textbf{Embeddings} & \textbf{SST-2} & \textbf{QQP} & \textbf{RTE} \\
  \midrule
  \multirow{3}{*}{RoBERTa} & RoBERTa & \textbf{0.953} & \textbf{0.919} & \textbf{0.801}\\
  & DistilBERT & 0.951 & 0.906 & 0.762\\
  & GloVe & 0.948 & 0.908 & 0.767\\
  \midrule
  \multirow{3}{*}{DistilBERT} & DistilBERT & \textbf{0.915} & \textbf{0.907} & \textbf{0.652} \\
  & RoBERTa & 0.906 & 0.903 & 0.623 \\
  & GloVe & 0.909 & 0.903 & 0.632 \\
  \bottomrule
\end{tabular}
\caption{\label{switch-table} Predictive accuracy using \algname with different word embeddings.}
\end{table}

\subsection{Effect of Different Embeddings}
To test the efficacy of computing \algname using the extracted representations from the target LM, we perform an experiment where we use the removal indices computed with 1) the representation from a different language model (e.g. removing indices for fine-tuning RoBERTa using the optimal removal index identified using DistilBERT data representations), and 2) GloVe pre-trained word embeddings \citep{pennington-etal-2014-glove}, as a third-party representation repository. 

\paragraph{Results.} As shown in \autoref{switch-table},
while alternate embeddings can still lead to improvements over the full data, using the representation from the target LM is beneficial and consistently outperforms other embeddings. 
The results suggest that low value data is likely a combination of (i) inherently noisy data (e.g. mislabeled instances) and (ii) instances that are harmful to specific models due to different model architectures and pre-training strategies.

\section{Conclusion}\label{sec:conclusion} In this work, we propose \algname  to address the model and dataset constraints that currently contribute to a computational bottleneck when computing Shapley-based data value estimates. 

\section*{Limitations} While we demonstrate the efficacy of \algname empirically, the current work is limited in terms of theoretical analysis. For example, while we have good empirical performance with a linear SVM, additional analysis could determine if there are optimal ways to select an alternative simple model architecture for the source classifier depending on the target classifier or dataset. Additionally, while we found a strong correlation between number of sampling chains and performance when the subset size was $>2\%$ of the training data size, the lower subset size threshold to observe this correlation may be dataset dependent, which additional analysis could address.

\bibliography{references}
\bibliographystyle{acl_natbib}

\clearpage

\appendix
\section{Additional Experiment Details}\label{app:exp-details}
In this section, we include additional experiment setup details.

\subsection{Datasets}

\begin{table*}[t!]
\centering
\small
\begin{tabular}{llllrr}%
  \toprule
  \multirow{2}{*}{\textbf{Dataset}} & \multirow{2}{*}{\textbf{GLUE Task Category}} & \multirow{2}{*}{\textbf{Task}} & \multirow{2}{*}{\textbf{Metric}} & \multicolumn{2}{c}{\textbf{Data Split}} \\
  \cmidrule{5-6}
  & & & & Train & Dev \\
  \midrule
  \vspace{1pt}
  SST-2 & Single Sentence Tasks & Sentiment & Acc. & 67k & 1.8k \\
  \vspace{1pt}
  QQP & Similarity and Paraphrase Tasks & Paraphrase & Acc./F1 & 364k & 40.4k \\
  \vspace{1pt}
  RTE & Inference Tasks & NLI & Acc. & 2.5k & 277 \\
  \bottomrule
\end{tabular}
\caption{\label{dataset-table-full} Statistics for each dataset. We use the train and development data splits as GLUE tasks have held out test set labels.}
\end{table*}

Dataset statistics are provided in \autoref{dataset-table-full}, with further description provided below.

\paragraph{SST-2:} Stanford Sentiment Treebank \citep{socher2013recursive} is a collection of English movie reviews with human annotations of their sentiment. The model is tasked with predicting a review's sentiment as positive or negative.

\paragraph{QQP: } Quora Question Pairs  \citep{qqpdataset} is a collection of English question pairs from the website Quora where the task is to determine if a pair of questions are similar in meaning.

\paragraph{RTE: } Recognizing Textual Entailment \citep{dagan2006pascal} combines several English datasets from annual textual entailment challenges, where the task is to predict if the \emph{text} entails the \emph{hypothesis} or not. 

\begin{table*}[t!]
\centering
\small
\begin{tabular}{llllllll}%
  \toprule
  \multirow{2}{*}{\textbf{Model}} & \multirow{2}{*}{\textbf{Method}} & \multicolumn{2}{c}{\textbf{SST-2}} & \multicolumn{2}{c}{\textbf{QQP}} & \multicolumn{2}{c}{\textbf{RTE}} \\
  \cmidrule{3-8}
   & & BS & LR & BS & LR & BS & LR\\
  \midrule
  \multirow{5}{*}{RoBERTa} & Full Dataset & 16 & $10^{-5}$ & 32 & $3 \times 10^{-5}$ & 16 & $3 \times 10^{-5}$\\
  & Leave-One-Out & 32 & $10^{-5}$ & -- & -- & 16 & $3 \times 10^{-5}$\\
  & KNN Shapley & 16 & $10^{-5}$ & 32 & $3 \times 10^{-5}$ & 16 & $3 \times 10^{-5}$ \\
  & Random & 32 & $3 \times 10^{-5}$ & 32 & $3 \times 10^{-5}$ & 16 & $3 \times 10^{-5}$ \\
  & \algname & 32 & $10^{-5}$ & 32 & $3 \times 10^{-5}$ & 16 & $3 \times 10^{-5}$\\
  \midrule
  \multirow{5}{*}{DistilBERT} & Full Dataset & 16 & $10^{-5}$ & 32 & $3 \times 10^{-5}$ & 32 & $3 \times 10^{-5}$\\
  & Leave-One-Out & 32 & $10^{-5}$ & -- & -- & 16 & $10^{-5}$\\
  & KNN Shapley & 16 & $10^{-5}$ & 32 & $3 \times 10^{-5}$ & 16 & $10^{-5}$\\
  & Random & 32 & $3 \times 10^{-5}$ & 16 & $3 \times 10^{-5}$ & 16 & $3 \times 10^{-5}$\\
  & \algname & 16 & $3 \times 10^{-5}$ & 16 & $10^{-5}$ & 16 & $3 \times 10^{-5}$\\

  \bottomrule
\end{tabular}
\caption{\label{selection-table1} Batch size (BS) and learning rate (LR) for the data selection experiment based on the hyperparameter sweep defined in \autoref{sec:exp}.}
\end{table*}

\begin{table*}[t!]
\centering
\small
\begin{tabular}{llllllll}%
  \toprule
  \multirow{2}{*}{\textbf{Model}} & \multirow{2}{*}{\textbf{Embeddings}} & \multicolumn{2}{c}{\textbf{SST-2}} & \multicolumn{2}{c}{\textbf{QQP}} & \multicolumn{2}{c}{\textbf{RTE}} \\
  \cmidrule{3-8}
   & & BS & LR & BS & LR & BS & LR\\
  \midrule
  \multirow{3}{*}{RoBERTa} & RoBERTa & 32 & $10^{-5}$ & 32 & $3 \times 10^{-5}$ & 16 & $3 \times 10^{-5}$\\
  & DistilBERT & 16 & $10^{-5}$ & 32 & $10^{-5}$ & 16 & $3 \times 10^{-5}$\\
  & GloVe & 16 & $3 \times 10^{-5}$ & 32 & $3 \times 10^{-5}$ & 32 & $3 \times 10^{-5}$\\
  \midrule
  \multirow{3}{*}{DistilBERT} & DistilBERT & 16 & $10^{-5}$ & 16 & $10^{-5}$ & 16 & $3 \times 10^{-5}$\\
  & RoBERTa & 32 & $10^{-5}$ & 32 & $10^{-5}$ & 32 & $10^{-5}$\\
  & GloVe & 32 & $10^{-5}$ & 32 & $3 \times 10^{-5}$ & 32 & $3 \times 10^{-5}$\\
  \bottomrule
\end{tabular}
\caption{\label{selection-table2} Batch size (BS) and learning rate (LR) for the embeddings switch experiment based on the hyperparameter sweep defined in \autoref{sec:exp}.}
\end{table*}

\subsection{Hyperparameters} For each experiment, we consider a limited hyperparameter sweep for each model, selection method, and task, with batch size $\in \{16, 32\}$ and learning rate $\in \{10^{-5}, 3\times 10^{-5}\}$. The rest of the hyperparameters are kept consistent across experiment conditions. We report the mean development set accuracy from five random initializations for which we fine-tune for 10 epochs and select the model checkpoint with the highest development set accuracy. Results from each hyperparameter sweep are reported in \autoref{selection-table1} and \autoref{selection-table2}.

\section{Additional Results}\label{app:results}

\begin{figure*}[t!]
\includegraphics[width=\linewidth]{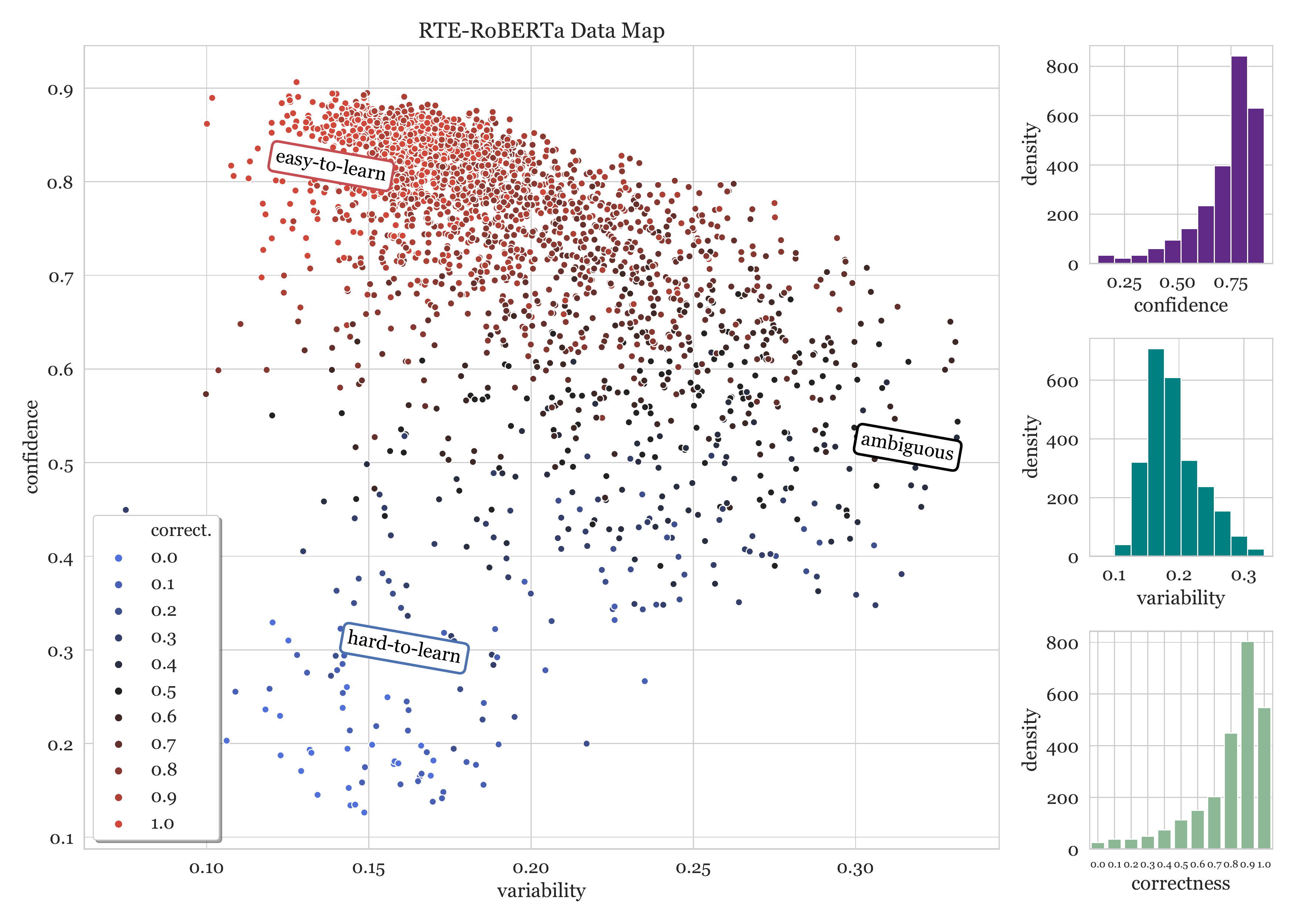}
\caption{\label{fig:map-roberta} Data map for RoBERTa trained on the RTE dataset.}
\end{figure*}
        
\begin{figure*}[t!]
\includegraphics[width=\linewidth]{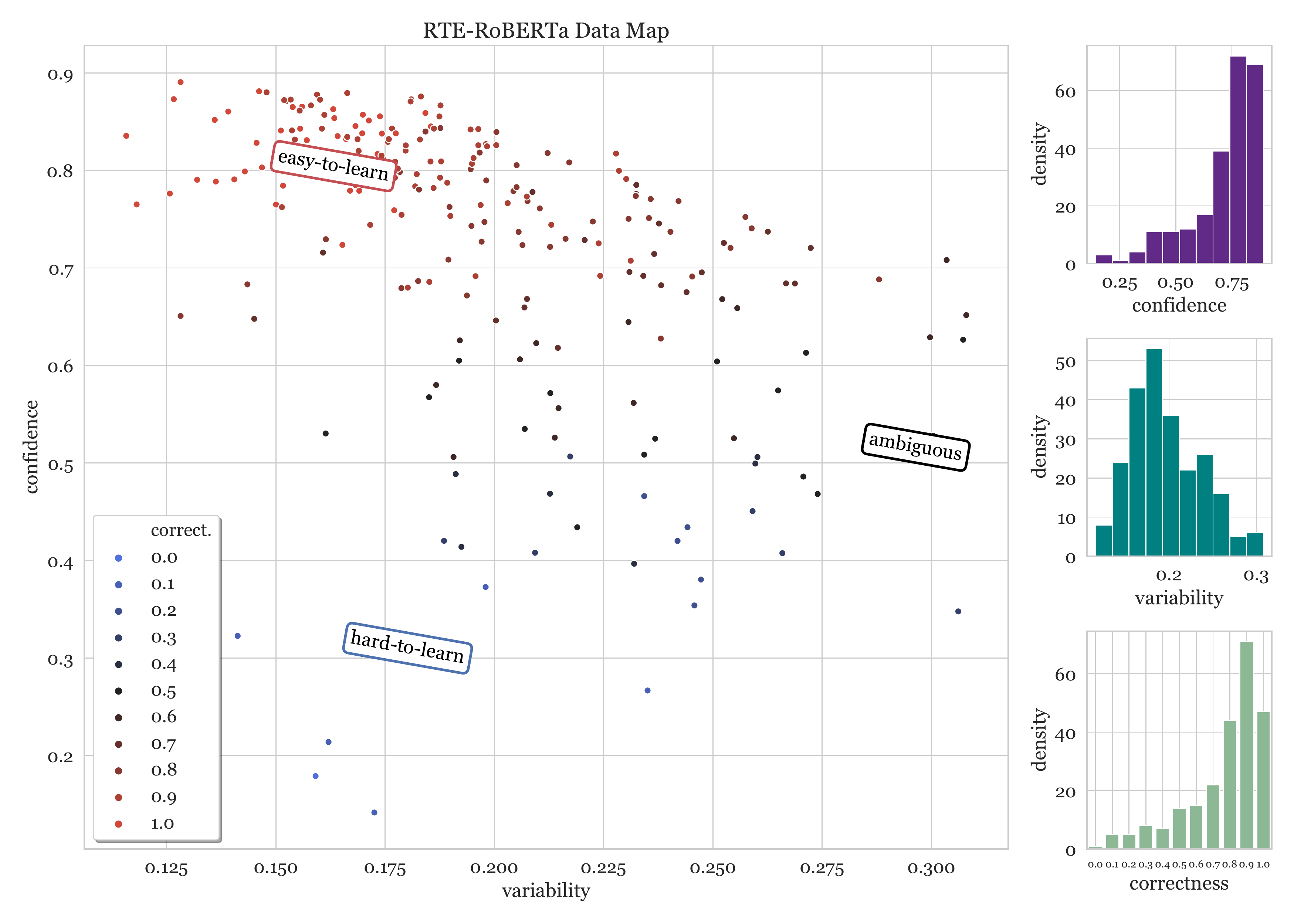}
\caption{\label{fig:map-roberta-combined} Data map showing location of training instances that were removed by \algname for RoBERTa on RTE.}
\end{figure*}

\subsection{Additional Data Selection Analysis}
While we compare directly with baseline selection methods that directly measure estimated data contribution, we perform an additional analysis by comparing the indices removed with \algname with the mapped training dynamics using data maps \citep{swayamdipta2020dataset}. Specifically, we first plot the data map for RoBERTa trained on RTE using the same hyperparameters as in \autoref{subsec:data-selection-exp}. Then, we plot the same data map showing only the data points that were identified by \algname to be harmful, i.e. removed from the fine-tuning training data. These are shown in \autoref{fig:map-roberta} and \autoref{fig:map-roberta-combined}, respectively.

We observe that a handful of instances in the hard-to-learn region (identified by \citet{swayamdipta2020dataset} to contain some mislabeled examples) were removed, as well as a small number of instances in the ambiguous region. Interestingly though, we observe that 1) most of the data points in RTE belonged to the easy-to-learn region, and 2) a cluster of easy-to-learn points were removed. \citet{swayamdipta2020dataset} found that too many easy-to-learn instances could decrease both in-distribution and out-of-distribution performance and noted that determining how to select an optimal balance of easy-to-learn and ambiguous examples, particularly in low data settings, was an open problem. As \algname achieved a performance gain over the full dataset performance, these results suggest that \algname may be effective to potentially determine an optimal balance and address this problem. We leave further analysis of this to future work.

\begin{table*}[t!]
\centering
\small
\begin{tabular}{llllll}%
  \toprule
  \multirow{2}{*}{\textbf{Model}} & \multicolumn{5}{c}{\textbf{Subset Size ($\%, \#$)}} \\
  \cmidrule{2-6}
  & 1 (25) & 2 (50) & 5 (125) & 10 (249) & 15 (374)\\
  \midrule
  RoBERTa & 0.119 & 0.013 & 0.892 & 0.929 & 0.942\\
  \vspace{1pt}
  DistilBERT & 0.240 & 0.104 & 0.613 & 0.776 & 0.714\\
  \bottomrule
\end{tabular}
\caption{\label{hp-analysis-nc} Correlations between number of chains and performance for each subset size on the RTE dataset.}
\end{table*}

\begin{table*}[t!]
\centering
\small
\begin{tabular}{lllllll}%
  \toprule
  \multirow{2}{*}{\textbf{Model}} & \multicolumn{6}{c}{\textbf{Number of Sampling Chains}} \\
  \cmidrule{2-7}
  & 2 & 5 & 10 & 15 & 20 & 25\\
  \midrule
  RoBERTa & -0.463 & 0.127 & -0.474 & 0.013 & 0.472 & 0.763 \\
  \vspace{1pt}
  DistilBERT & 0.027 & -0.034 & 0.530 & 0.447 & 0.737 & 0.692\\
  \bottomrule
\end{tabular}
\caption{\label{hp-analysis-ss} Correlations between subset size and performance for each number of sampling chains on the RTE dataset.}
\end{table*}

\subsection{Sampling Hyperparameter Analysis} Pearson's correlation coefficients for the sampling parameter analysis in \autoref{sec:exp} are reported in \autoref{hp-analysis-nc} and \autoref{hp-analysis-ss}, where each result represents the mean of five sampling and chain computation trials.
\end{document}